\pdfoutput=1

\documentclass[11pt]{article}

\usepackage[preprint]{acl}

\usepackage{times}
\usepackage{latexsym}

\usepackage[T1]{fontenc}

\usepackage[utf8]{inputenc}

\usepackage{microtype}

\usepackage{inconsolata}

\usepackage{graphicx}
\usepackage{enumitem}

\usepackage{algorithm}
\usepackage{algpseudocode}
\usepackage{algorithmicx}
\usepackage{tabularx}   
\usepackage{booktabs}   
\usepackage{xcolor}     
\usepackage{tcolorbox}
\usepackage{float}

\usepackage{array} 

\newcolumntype{P}[1]{>{\centering\arraybackslash}p{#1}}

\newcommand{\name}{Comp-LLM}

\title{Experts are all you need: A Composable Framework for Large Language Model Inference}

\author{
  Shrihari Sridharan \quad Sourjya Roy \quad Anand Raghunathan \quad Kaushik Roy  \\
  School of Electrical and Computer Engineering \\
  Purdue University \\
  \texttt{\{sridhar4, roy48, araghu, kaushik\}@purdue.edu}
}

\begin{document}
\maketitle
\begin{abstract}
Large Language Models (LLMs) have achieved state-of-the-art accuracies in a variety of natural language processing (NLP) tasks. However, this success comes at the cost of increased model sizes which leads to additional computational burden. Mixture of Experts (MoEs) overcome this bottleneck by decoupling model capacity from computation by only activating a subset of parameters or "experts". However, these models require joint pretraining of these experts along with router and do not model multi-step reasoning. In contrast, multi-agent frameworks improve reasoning by decomposing complex problems into modular subtasks. However, these frameworks rely on sequential “plan–act–observe" loops, which introduce significant latency. Our work, ~\name{}, addresses these challenges by introducing a composable inference framework that enables cross-expert collaboration via an explicit sub-query dependency graph. Comp-LLM consists of three components: (1) A Sub-query Generator that decomposes an input query, assigns each sub-query to an appropriate expert using embedding similarity, and constructs a dependency graph; (2) A Query Executor that processes nodes in the graph and identifies opportunities for parallelism based on dependencies and resource constraints; and (3) A Response Aggregator that synthesizes intermediate expert responses into a coherent final answer. Across several benchmarks, Comp-LLM achieves up to 11.01\% accuracy improvement over monolithic LLMs of similar size, while offering 1.67$\times$--3.56$\times$ reduction in model size with no significant degradation relative to the largest model in its family. Additionally, Comp-LLM provides 1.1$\times$--1.7$\times$ latency improvement compared to sequential sub-query processing.

\end{abstract}

\section{Introduction}
\label{sec:introduction}
\begin{figure}[htb]
  \includegraphics[width=\columnwidth]{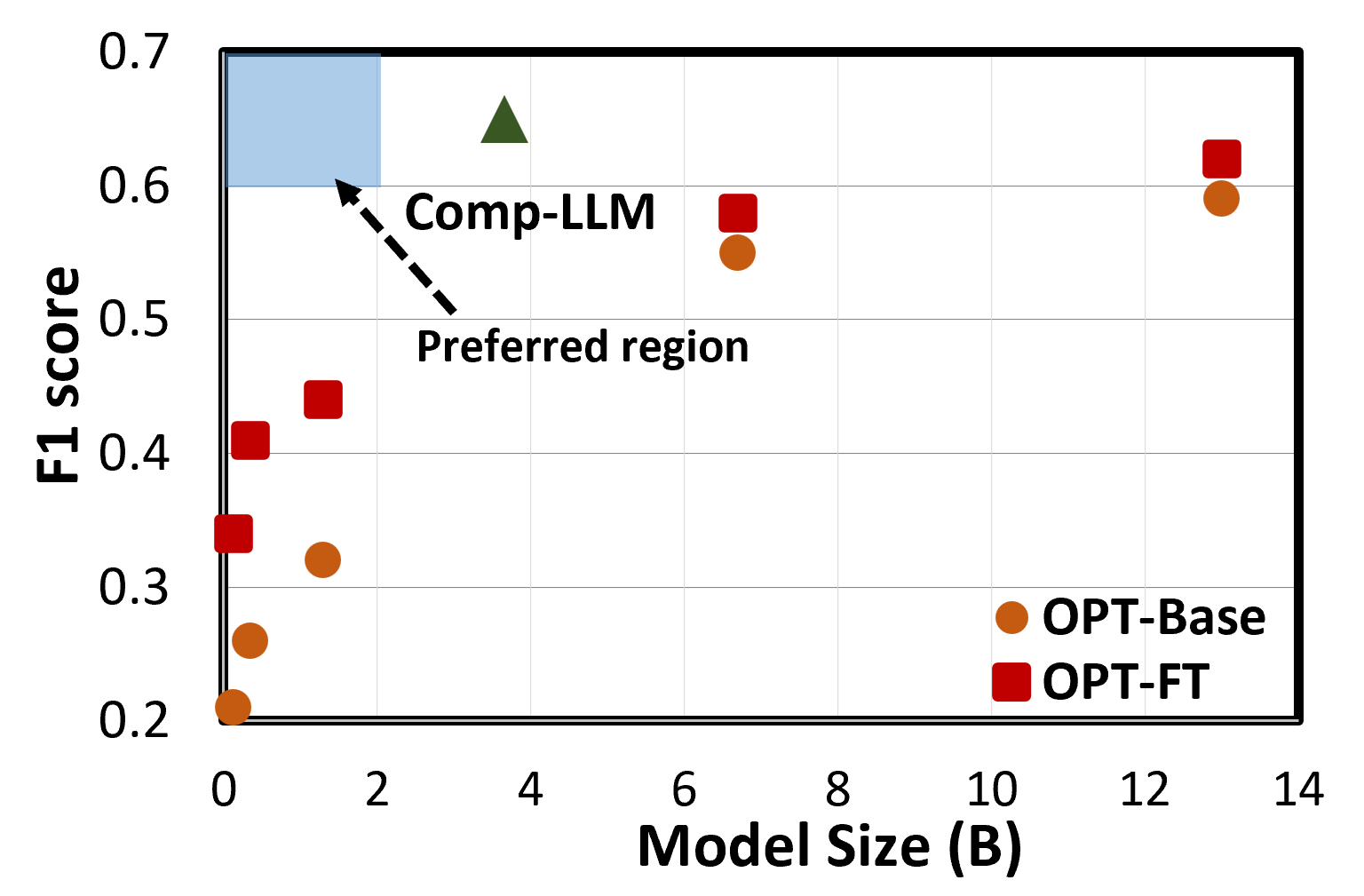}
  \caption{F1 score vs Model Size for different configurations of OPT-base and OPT-FT (finetuned)~\citep{opt} on Expert-QA-P (2 experts) benchmark. The proposed ~\name{} (3.65B) produces higher F-1 score compared to OPT-Base (13B) and OPT-FT (13B). }
  \label{fig:modelacc}
\end{figure}
Large Language Models (LLMs) such as GPT~\citep{gpt3,openai2024}, Claude~\citep{claude} and Llama 3~\citep{llama3} have demonstrated unprecedented success in a variety of natural language processing (NLP) tasks including text generation, language translation, and code generation~\citep{llmsurvey1, llmsurvey2}. These models are pre-trained on a massive text corpora and subsequently finetuned for downstream applications. These impressive accuracy gains, however, come at the cost of significant increase in model size, which demand extensive data and substantial computational resources, as demonstrated by the empirical scaling laws of LLMs.~\citep{scalelaws}. For example, on the AlpacaEval 2.0 benchmark~\citep{alpaca2.0}, larger models consistently obtain higher accuracies compared to their smaller counterparts.


To that end, Mixture of Experts (MoEs) has been proposed to improve parameter efficiency by routing input tokens to specialized sub-networks (experts) using a router. However, this approach has two major limitations. First, the experts and the gating router require expensive joint pre-training, and the router must be retrained to add new experts. Second, token-level routing lacks an explicit model of logical dependencies, limiting its ability to coordinate among complex reasoning steps. Agentic frameworks, in contrast, decompose complex problems into sub-tasks and allow multiple specialized agents to collaborate through planning and communication~\citep{yao2023react,chen2024agentbench,wang2023voyager}. While this improves the reasoning capabilities, these systems are typically dynamic, relying on sequential “plan–act–observe’’ loops to iteratively construct solutions. This sequential interaction between agents leads to high latency and is suboptimal for the large class of static reasoning tasks (e.g., multi-hop question answering) where the full execution plan is known ahead of time but cannot be exploited for parallelism.


To address the aforementioned challenges, we propose ~\name, a composable LLM inference framework that improves question-answering and reasoning capabilities while reducing memory footprint through sub-query generation and cross-expert collaboration. ~\name{} consists of three key components -- a Sub-query Generator, a Query Executor and a Response Aggregator. The Sub-query Generator begins by decomposing the complex query into simpler sub-queries and identifies pairwise dependencies among them. It subsequently routes each sub-query to the most appropriate expert based on similarity between it's embeddings and the pre-computed expert embeddings. A dependency graph of the sub-queries is then generated to preserve logical consistency in answering the original query. Next, each sub-query in the dependency graph is processed in topological order by the Query Executor, which consists of experts fine-tuned on their specific domain data. The responses generated at each node provide context for the subsequent dependent nodes in the graph.

While individually answering sub-queries improves the reasoning capabilities of ~\name{}, it increases the overall latency due to processing multiple sub-queries. Therefore, the Query Executor consists of a runtime scheduler that determines an execution plan by identifying the nodes within the dependency graph that can execute in parallel, thereby minimizing overall latency. Finally, the Response Aggregator takes in the original query and the responses from the experts to generate a final response.

We perform a comprehensive evaluation of \name{} on the  MultiExpertQA-P and MultiExpertQA-All benchmarks, which consists of queries with no sub-query dependencies and queries with dependencies, respectively, across different expert domains from existing benchmarks. Our results demonstrate that \name{} achieves 1.67x-3.56x reduction in model size with comparable accuracy as the largest model in its family. Additionally, we also observe 11.01\% accuracy improvement on average in comparison with a model of similar size. Finally, \name{} also achieves 1.1x-1.7x improvement in latency over the sequential processing of sub-queries. Figure ~\ref{fig:modelacc} illustrates the comparison of F1 scores and model sizes between various base OPT models and those that have been fine-tuned on all the expert datasets.


\section{Related Work}
\label{sec:relatedwork}

In this section, we describe the prior works related to ~\name{} and place our approach in their context.

\noindent\textbf{Mixture of Experts.} Mixture of Experts (MoE)~\citep{switchtr,mixtral} is a monolithic model which consists of  several specialized sub-networks (semantic experts) with a gating function to dynamically route input tokens. However, MoEs require joint pretraining of the experts and gating function which necessitates excessive computational resources. In constrast, ~\name{} utilizes independently pretrained domain expert LLMs to process an input query.

\noindent\textbf{Model Fusion.} Several works have explored combining different LLMs to enhance performance. Model fusion approaches can be divided into two main categories: weight merging and model ensembling. Weight merging combines the parameters of multiple LLMs into a single unified model. For instance, ~\citet{choshen2022fuse} and ~\citet{fisher} fuse models using simple or weighted averaging of parameters. Other works such as ~\citet{zhang2023composing} and ~\citet{huang2024lorahub} integrate model adapters using arithmetic operations for improved generalization. However, weight merging requires the individual models to share a common architecture and train from the same random initialization.

Model ensembling methods (Jiang et al., 2023; Huang et al., 2024b) combine outputs from multiple pretrained LLMs to produce accurate responses. However, when the outputs are aggregated, there is information loss since information from all the individual models maybe not be captured accurately.  To that end, adaptive routers (Ong et al., 2024; Lee et al., 2024; Stripelis et al., 2024; Srivatsa et al., 2024) overcome some of these limitations by directing inputs to the most relevant expert. However, routing only to one expert might be suboptimal for real-world tasks that require multiple reasoning steps. Moreover, these methods require retraining the router whenever a new expert is augmented, thereby increasing computational costs for training. In contrast, CompLLM breaks down input queries into sub-queries, routing each sub-query to the most suitable expert, enabling effective collaboration among multiple experts without any retraining.

\noindent\textbf{Decomposed Prompting.} Various prompting techniques ~\citep{cot,cotsc,psprompt} and adaptive reasoning frameworks ~\citep{decomp,adapt} have emerged as powerful methods for guiding LLMs to generate intermediate reasoning steps. ~\name{}, in addition to decomposing the query to sub-queries, also generates a dependency graph that allows for parallel execution of sub-queries based on data dependencies and resource constraints. Each sub-query is then directed to a specialized expert leading to collaboration among these experts. 
 
\noindent\textbf{Compression Techniques.} Various algorithmic techniques such as quantization~\citep{gptint8,outquant,awq}, pruning~\citep{ebert,llmpruner} and knowledge distillation~\citep{pkdistill,contrastkd,metakd} have been proposed to reduce model size. These works complement our approach and can be integrated into ~\name{} to achieve further performance improvements.

\noindent\textbf{Agentic and Multi-Agent Reasoning Frameworks.}
In order to produce a coherent answer, agentic frameworks decompose complex queries into various sub-queries which are then individually answered by planning, tool invocation or inter-agent communication (Yao et al., 2023a; Chen et al., 2024; Wang et al., 2023a). Some of these approaches include ReAct (Yao et al., 2023a), ReWOO (Hong et al., 2024), Tree-of-Thoughts (Yao et al., 2023b), Graph-of-Thoughts (Besta et al., 2024), CAMEL (Li et al., 2023b), AutoGen (Wu et al., 2023), and debate-based multi-agent systems (Du et al., 2023). These approaches are suited for dynamic and interactive tasks, where the next action is based on the current response. Therefore, they operate through plan-act-observe loops across multiple iterations. However, for static reasoning tasks where the full logical dependencies between the sub-queries can be determined in advance, this approach leads to increased latency.
In contrast, \name{} explicitly constructs a dependency graph over sub-queries and identifies opportunities for parallel execution across pretrained experts.

\vspace{-8pt}
\section{~\name{}: A Framework for Composing Pre-trained LLMs}
\label{sec:framework}

\begin{figure*}[htb]
  \includegraphics[width=\textwidth]{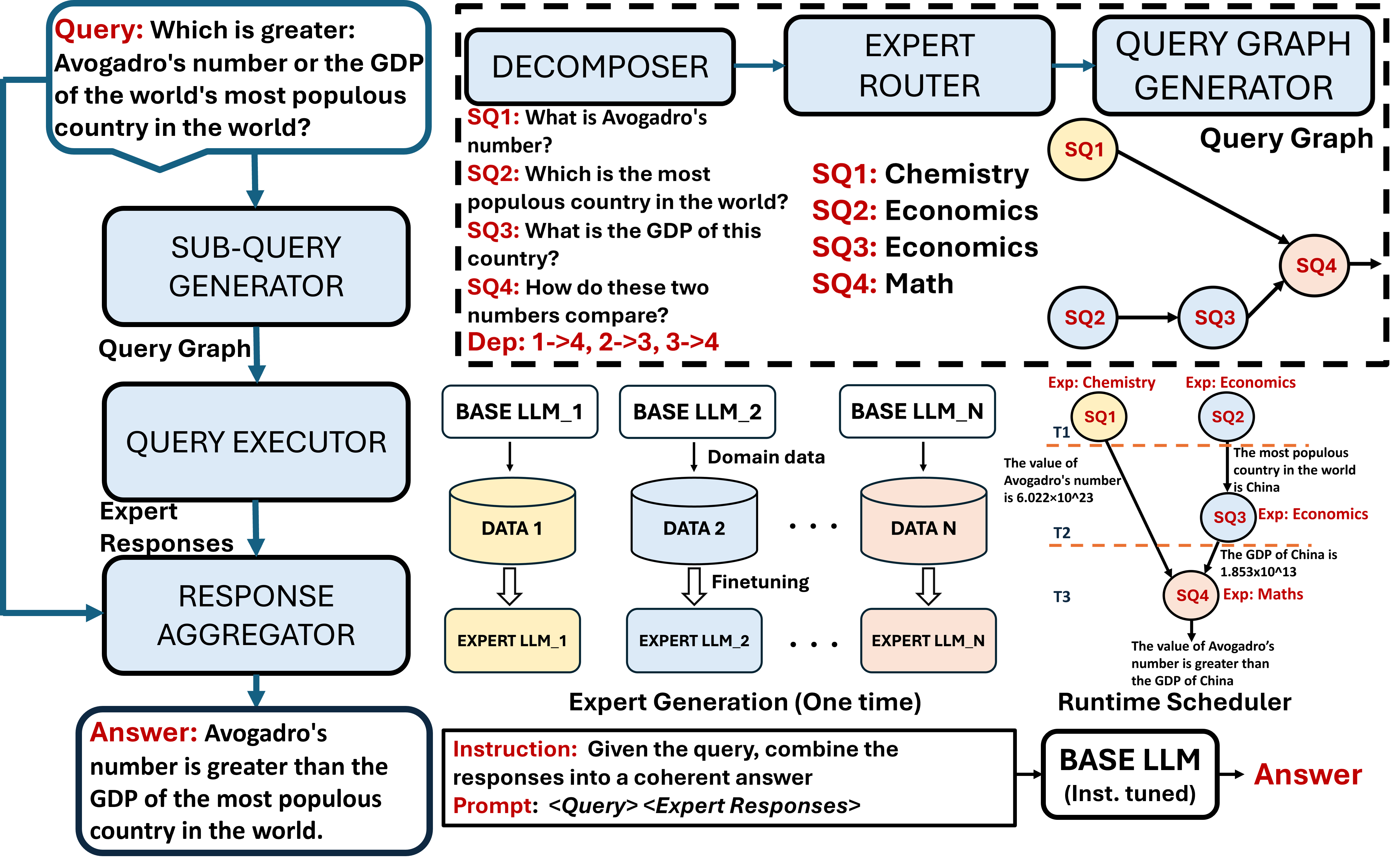}
  \caption{~\name{} framework improves reasoning capabilites of LLMs. It consists of three key components: Sub-query Generator, Query Executor and Response Aggregator}
  \label{fig:overall}
\end{figure*}

~\name{} is a composable framework designed to enhance the reasoning capabilities of LLMs while significantly reducing their memory footprint. Figure ~\ref{fig:overall} presents an overview of the ~\name{} framework. Given an input query, the Sub-query Generator first generates a query graph of sub-queries that help answer the original query by considering the dependencies between them. The Query Executor then executes each query in the graph and incorporates a runtime scheduler to identify sub-queries that can be executed in parallel. Finally, the Response Aggregator synthesizes the individual sub-query responses into a coherent overall response. The following subsections will now describe each component in detail.
\subsection{Sub-query Generator}
The Sub-query Generator consists of a three stage pipeline to produce a query graph. First, the decomposer breaks down the input query into different sub-queries and generates the pairwise dependencies between them. Next, the expert router routes each sub-query to an expert based on the similarity between the query embedding and the pre-computed expert embeddings. Finally, the query graph generator converts the pairwise dependencies and the assigned experts into a query graph.
\subsubsection{Decomposer}
\label{sec:decomp}

\begin{figure}[htb]
  \includegraphics[width=\columnwidth]{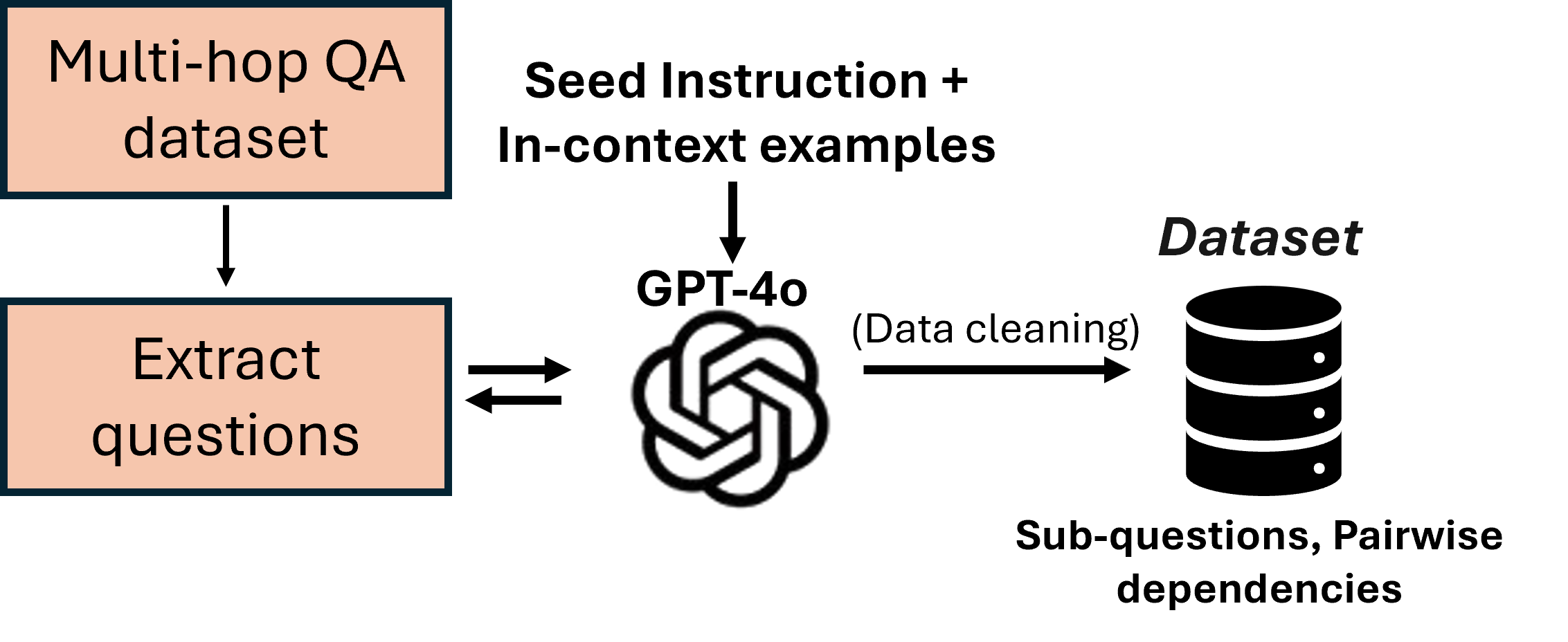}
  \caption{Sub-query Generator dataset generation}
  \label{fig:subqgen}
  \vspace{-12pt}
\end{figure}

The Decomposer analyzes the input query and identifies relevant sub-queries that assist in answering the original query. In order to maintain logical consistency, it is important to identify the dependencies between these sub-queries. We finetune the pretrained LLM through supervised finetuning to output these sub-queries and their dependencies. To that end, we created a dataset consisting of queries, sub-queries and their pairwise dependencies. We utilized existing multi-hop question answering datasets, such as HotpotQA~\citep{hotpotqa} and MuSiQue~\citep{musique} and first selected a subset of queries from them. Next, we present these queries to GPT-4o~\citep{openai2024} along with few in-context examples and a seed instruction (see Appendix ~\ref{sec:appendix1} for prompt details). For datasets that already provide query decomposition, like MuSiQue, we utilize the existing decomposition and employ GPT-4o to only identify the dependencies between the sub-queries. In cases where the decomposition is not provided, GPT-4o handles both decomposing the queries into sub-queries and determining the dependencies between them. We represent the pairwise dependencies as $SQ\_i \rightarrow SQ\_j$, where $SQ\_j$ can be answered only after $SQ\_i$ is completed. We then parse the output and concatenate the sub-queries and dependencies into a target sequence. In order to delineate between different parts of the target sequence, we introduce special tokens: \textbf{<dep>} and \textbf{<\textbackslash{dep}>} around the dependency string, and \textbf{<q>} and \textbf{<\textbackslash{q}>} around each sub-query. 

\begin{figure}[htb]
  \centering
  \includegraphics[scale=0.4]{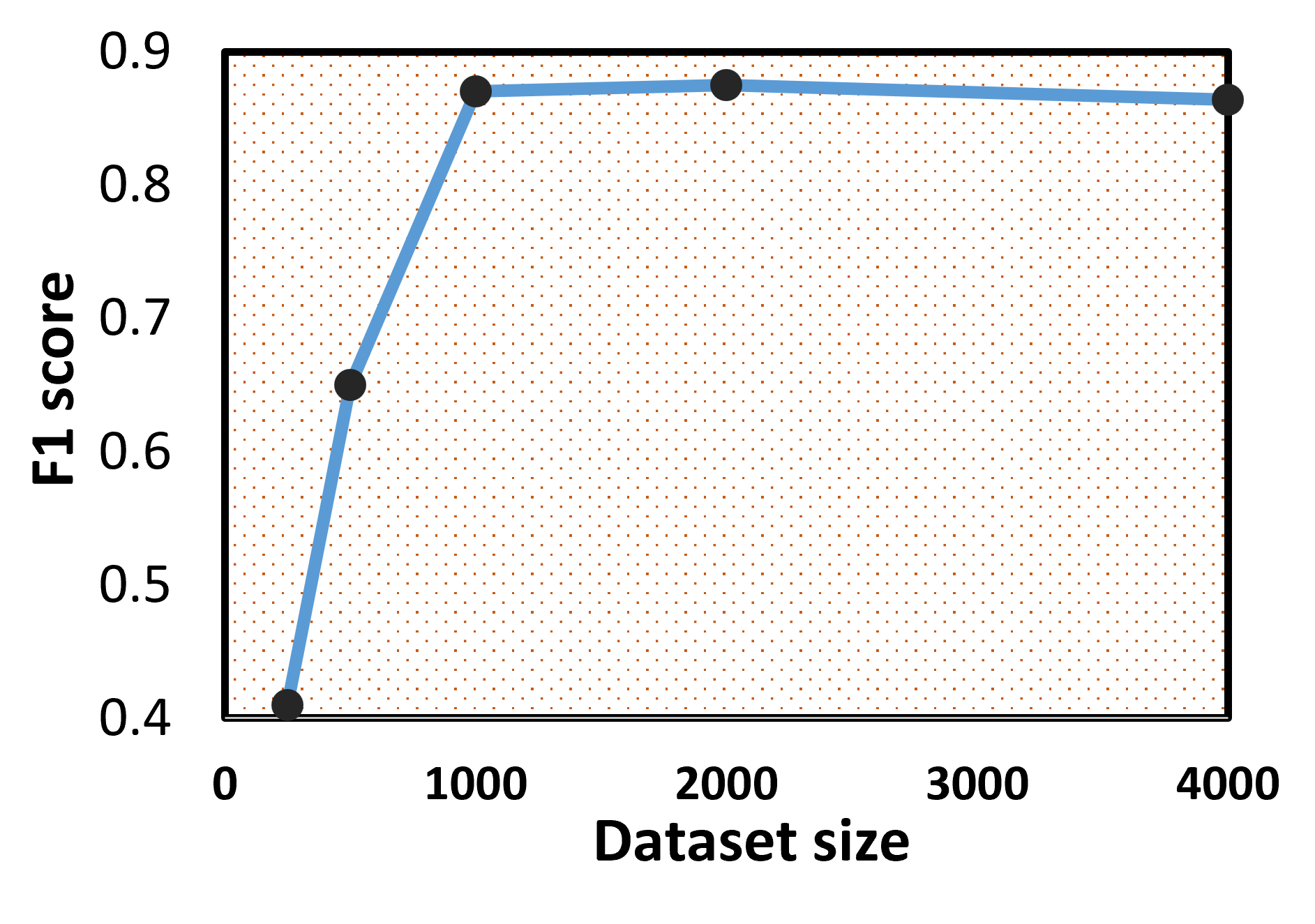}
  \caption{F1 score varying the training dataset size for Sub-query Generator.}
  \label{fig:subqres}
\end{figure}

We compare our Sub-query Generator with a base model prompted with few in-context learning (ICL) examples for Llama 2 7B~\citep{llama2} and observe that our Sub-query Generator significantly outperforms few-shot prompting on both the MuSiQue and HotpotQA datasets, as indicated in Table ~\ref{tab:decomp}. To identify the optimal number of training examples for the Sub-query Generator, we conducted an ablation study, as illustrated in Figure ~\ref{fig:subqres}, where we evaluated test performance across varying training dataset sizes. The results show that the F-1 score plateaus or slightly decreases beyond 1000 examples. Therefore, we choose to train our Sub-query Generator with 1000 examples.

\begin{table}
  \centering
  \footnotesize
     \begin{tabular}{ccP{2cm}}
         \hline
         Dataset & Llama 2+ICL & Sub-query Generator \\
         \hline
         HotpotQA & 0.48 & 0.87 \\
         MuSiQue & 0.42 & 0.86 \\
         \hline
     \end{tabular}
     \captionof{table}{Comparison of Sub-query Generator accuracy and Llama 2 few-shot prompting with different multi-hop datasets}
     \label{tab:decomp}
     \vspace*{-12pt}
\end{table}


\subsubsection{Expert Router}

\begin{figure}[htb]
  \includegraphics[width=\columnwidth]{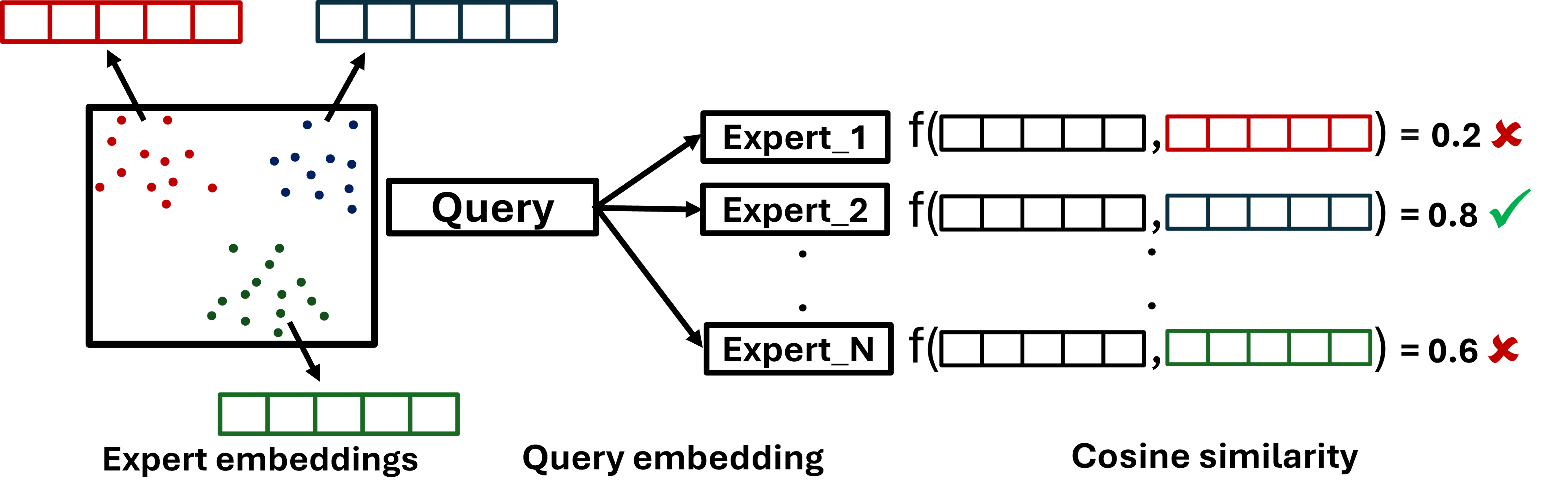}
  \caption{Expert Router}
  \label{fig:router}
  \vspace*{-12pt}
\end{figure}

After the decomposer splits the query into sub-queries, the expert router assigns each sub-query to an expert based on its similarity to the expert's training data distribution. For each query in the dataset, we compute the contextual embeddings using the approach proposed in ~\citep{sbert}, \emph{viz.}, last token hidden state (LTHS) and mean pooling (MP), and modify it for decoder-based LLMs. Unlike encoder-based approaches that rely on the \texttt{[CLS]} token for the embedding representation, LTHS employs the final token of the sentence as a proxy instead~\citep{ood}. This is because in decoder-based LLMs, each token is generated based on the preceding context, making the final token an aggregate representation of the sentence. The second technique, mean pooling, computes the average across  the embeddings of all tokens in an input sequence, thereby capturing a broader range of semantic information. After we obtain the embeddings for each sentence, we average these representations to create a single embedding that represents the expert dataset. We note that this is a one-time procedure performed before inference. 

During inference, for each sub-query, we compute it's embeddings using the previously described techniques and identify the expert with the highest cosine similarity score. However, there is a chance that the sub-query is not relevant to any of the experts. In such cases, choosing the expert with the highest cosine similarity score is incorrect.  To address this issue, we introduce a similarity threshold, denoted as \textit{sq\_sim} which the previously computed cosine similarity score must exceed. The optimal value of \textit{sq\_sim} is emperically chosen based on error rate observation. If the similarity score for a sub-query is less than \textit{sq\_sim}, the sub-query is routed to a base LLM that isn't domain-specific. As shown in Table ~\ref{tab:experts_performance}, our experiments reveal that MP provides a lower error rate (i.e.) percentage of incorrectly routed sub-queries compared to LTHS across various expert configurations. This is because LTHS does not capture accurate global context since it is biased towards the final tokens, which becomes more pronounced as the sequence length increases. Consequently, we implement mean pooling (MP) approach for the expert router. We also perform an experiment where, instead of averaging the entire dataset as one single embedding, we create multiple embeddings by splitting the dataset into subsets and compute the representation for each of the subsets. For a two-expert configuration for Llama 2, as shown in Table ~\ref{tab:embeddings_performance}, we observe that the error rate goes down for both the techniques as the number of embeddings increases. However, this escalates the computational complexity, since the new sub-query needs to be compared against multiple embeddings within the same dataset. Therefore, we choose the number of embeddings to be 1 in our experiments as it offers the best tradeoff between accuracy and computational complexity. We also note that while other methods that utilize classifiers or LLMs as routers tend to be more accurate~\citep{sbert}, these approaches are excessive for our needs. Since the Decomposer breaks down the query into sub-queries which has simpler semantics, our technique effectively addresses the requirements without the added complexity.

\vspace{-10pt} 

\begin{table}[h]
  \centering
  \footnotesize 
  \begin{tabular}{ccc}
    \hline
    \textbf{\#Experts} & \textbf{MP} & \textbf{LTHS} \\
    \hline
    2     & 0.8  & 2.3 \\
    3     & 1.1  & 3.2 \\
    4     & 1.2  & 3.6 \\
    \hline
  \end{tabular}
  \caption{Error rate for MP and LTHS methods by varying number of experts.}
  \label{tab:experts_performance}
\end{table}

\vspace{-15pt} 

\begin{table}[h]
  \centering
  \footnotesize 
  \begin{tabular}{ccc}
    \hline
    \textbf{\#Embeddings} & \textbf{MP} & \textbf{LTHS} \\
    \hline
    2     & 0.9  & 2.3 \\
    3     & 0.7  & 2.1 \\
    4     & 0.5  & 1.8 \\
    \hline
  \end{tabular}
  \caption{Error rate for MP and LTHS methods by varying number of embeddings for the expert datasets.}
  \label{tab:embeddings_performance}
\end{table}
\subsubsection{Query Graph Generator}
To represent the pairwise dependencies derived from the decomposer, we construct a query graph as a directed acyclic graph (DAG). Each node in the query graph contains two attributes: the sub-query itself and the designated expert. We then iterate over all pairwise dependencies and build the query graph, where each directed edge signifies a specific dependency between nodes. As a result, each node in the query graph can have multiple incoming and outgoing edges. This query graph, therefore, captures the structure and dependencies of the decomposed sub-queries while assigning each sub-query to the appropriate expert for processing.
~\vspace{-10pt}
\subsection{Query Executor}

After the Sub-query Generator produces a query graph, the Query Executor processes each node using its assigned expert. The resulting responses are then utilized as context for all dependent nodes in the query graph. The Query Executor consists of experts that are finetuned from base LLMs using domain specific data. A straightforward approach to processing the nodes in the query graph would be to process them sequentially. Consequently, this leads to a significant increase in latency, especially for complex queries with a large number of nodes. To that end, we propose a runtime scheduler that employs a scheduling algorithm to identify opportunities for parallelism within the query graph, considering data dependencies and resource constraints. 
The algorithm for scheduling nodes in the query graph QG is described in Algorithm ~\ref{alg:schedule}. Given the query graph (QG) that consists of data dependencies of each node, the sub-query mapping that indicates the expert designated to process a specific node ($M\_{GPU}$), and the resource mapping that specifies the experts assigned to each resource ($M\_{SQ}$), our algorithm identifies opportunities for parallelizing the nodes within QG and executes the sub-queries. We also maintain a status flag ($S\_{GPU}$) for each GPU to indicate if its free (1) or busy (0).
We maintain a $ready\_list$ that stores the nodes which are ready for execution in QG. We initially populate this list with nodes whose parents have no dependencies (Lines 5-9). Based on the status of GPU resources, we schedule these nodes ($busy\_nodes$) from the $ready\_list$  (Line 11). These $busy\_nodes$ that are assigned to their hardware resource can now be processed in parallel (Lines 11-20). For each sub-node in $busy\_nodes$, the designated expert is identified and used to generate the expert response (Lines 14-15), and this response is appended to its child nodes as context for it's sub-query (Line 16). The sub-node is then marked as completed (Line 17). We also check if the parent nodes of the current sub-node are now ready for processing and add them to the $ready\_list$ (Lines 18-20). After parallel execution, the processed nodes are removed from the $ready\_list$ (Line 24) and we change the status of the GPU resources to free (1). This process continues iteratively until the $ready\_list$ is empty, with the final output consisting of responses from the leaf nodes in the graph.

\begin{algorithm}
\caption{Runtime Scheduler Algorithm}\label{alg:schedule}
\footnotesize 
\begin{algorithmic}[1]
   \State \textbf{Input:} Query graph $QG = (V, E)$, where $V = \{SQ_1, \dots, SQ_n\}$, $E = \{(SQ_i, SQ_j)\}$; HW resource mapping $M_{GPU} = \{Exp_i: GPU_i\}$; Sub-query mapping $M_{SQ} = \{SQ_i: Exp_i\}$; HW status $S_{GPU} = \{GPU_i: 1\}$
    \State \textbf{Output:} Leaf node Expert responses 
    \Procedure{Execute}{$QG$,$M_{GPU}$,$M_{SQ}$}
    \State ready\_list $\gets []$ 
    \For {each node in QG}
        \If {are\_parents\_completed(node)}
            \State ready\_list.append(node)
        \EndIf
    \EndFor
    \While {ready\_list is not empty}
        \State busy\_nodes = schedule\_nodes(ready\_list,$M_{GPU}$)
        \State \textcolor{blue}{// All busy nodes can now execute in parallel}
        \For {each sub\_node in parallel in busy\_nodes}
            \State Expert = $M_{SQ}$(sub\_node)
            \State response $\gets$ Expert(sub\_node)
            \State append\_context(response, child\_nodes)
            \State mark\_completed(sub\_node)
            \For {each par in parent\_nodes}
                \If {are\_parents\_completed(par)}
                    \State ready\_list.append(par)
                \EndIf
            \EndFor
        \EndFor
        \State ready\_list.remove(busy\_nodes)
    \EndWhile
    \EndProcedure
\end{algorithmic}
\end{algorithm}

~\vspace{-14pt}
\subsection{Response Aggregator}

As shown in Table ~\ref{tab:response_prompt} in Appendix ~\ref{sec:appendix1}, once we obtain the leaf expert responses, we prompt an instruction-finetuned base LLM to combine these responses to produce a final coherent response. For the input to the base LLM, we concatenate the original query with the leaf node responses. We focus solely on the leaf node responses because the other expert responses have already served as context for different nodes in the query graph. This approach ensures that we incorporate the most specific and relevant information from the expert graph while maintaining the original context of the query.

\section{Experimental Methodology}
\label{sec:experiments}

In this section, we describe the experimental setup and networks and benchmarks used to evaluate ~\name{}.

\noindent\textbf{Experimental Setup.} We implemented \name{} in PyTorch using the Huggingface library~\citep{hf} on a node equipped with 4 NVIDIA A100 GPUs, each with 80GB of memory. The Sub-query Generator was finetuned on a question answering task objective for 10 epochs with our constructed dataset, described in Section ~\ref{sec:decomp}.
The expert models were finetuned on the language modeling task using LoRA~\citep{lora} for 1 epoch. We also set $sq\_sim$=0.7 for the expert router. The details of the other hyperparameters and prompts used are reported in Appendix ~\ref{sec:appendix} and ~\ref{sec:appendix1} respectively. 

\noindent\textbf{Networks and Benchmarks.} For our base LLMs, we selected various Llama 2~\citep{llama2} models ranging from 7B to 70B parameters, as well as OPT~\citep{opt} models with sizes ranging from 125M and 13B parameters. We describe the configurations used in our framework for each model in Table ~\ref{tab:mdlconfig}. To evaluate ~\name{}, we adopted a similar method described in Section ~\ref{sec:decomp} and utilized datasets generated by ~\citep{camel} across three domains: Chemistry, Biology, and Math. We developed two benchmarks, MultiExpertQA-P and MultiExpertQA-All. MultiExpertQA-P consists of queries with independent sub-queries, while MultiExpertQA-All includes queries where sub-queries have dependencies. For MultiExpertQA-P, we selected queries from two distinct domains (datasets) and used the GPT-4o model to construct new queries by merging them. For MultiExpertQA-All, we first identified key entities in each question and generated a fact about the entity in a certain domain using GPT-4o and replaced these entities in the original question. Additionally, we also injected queries from unrelated domains in the benchmark. Each benchmark has two variants: one involving two expert domains and another with three domains. We augmented a base model in ~\name{} in case none of the experts are appropriate for answering a query.


\begin{table}[H]
  \centering
  \footnotesize
  \begin{tabularx}{\columnwidth}{XP{1.5cm}cP{2cm}} 
    \hline
    \textbf{Network} & \textbf{Sub-query Generator} & \textbf{Experts} & \textbf{Response Aggregator} \\
    \hline
    Llama 2  & 7B   & 7B   & 7B \\
    OPT     & 1.3B  & 350M  & 1.3B \\
    \hline
    
  \end{tabularx}
  \caption{Model configuration used for Llama 2 and OPT models in ~\name{}.}
  \label{tab:mdlconfig}
\end{table}
\section{Results}
\label{sec:results}

In this section, we report the accuracy and performance improvements of ~\name{} over state-of-the-art LLMs.

\subsection{Accuracy Results}

We evaluate ~\name{} on MultiExpertQA-P and MultiExpertQA-All benchmarks as shown in Table ~\ref{tab:overall}. These benchmarks consist of queries requiring two or three experts to generate responses. For Llama 2 base models, ~\name{} demonstrates higher accuracy compared to the 34B model, and when compared to the 70B model, it incurs no significant accuracy degradation while achieving 2x and 1.67x reduction in model size for two and three experts respectively. We also constructed a stronger baseline individually fine-tuning the Llama 2 models on the expert datasets, which resulted in improved accuracy over their base versions. Nevertheless, ~\name{} outperforms these fine-tuned models as well. Similarly, for OPT models, we observe that ~\name{} outperforms OPT-13B base and finetuned versions with 3.56x and 3x reduction in model size for two and three experts respectively. We also observe that on the MultiExpertQA-All benchmark, both the base and fine-tuned versions experience a drop in accuracy due to error propagation, where inaccuracies in earlier sub-query responses leads to more errors in the subsequent stages. In contrast, our framework demonstrates lower accuracy degradation on the MultiExpertQA-All benchmark, since each sub-query is handled by the most appropriate domain expert, and logical consistency is maintained across various sub-queries.


~\vspace{-15pt}
\begin{table*}[htb]
    \centering
    \caption{Accuracy results comparing ~\name{} to different state-of-the-art LLMs.}
    \small 
    \begin{tabularx}{\textwidth}{l*{1}{c}*{5}{X}*{6}{X}}  
        \toprule
        & & \multicolumn{5}{c}{\textbf{Llama 2}} & \multicolumn{6}{c}{\textbf{OPT}} \\
        \cmidrule(lr){3-7} \cmidrule(lr){8-13}  
        \textbf{Benchmark} & \textbf{\#Exp} & \textbf{7B} & \textbf{13B} & \textbf{34B} & \textbf{70B} & \textbf{Ours (35B)} & \textbf{125M} & \textbf{350M} & \textbf{1.3B} & \textbf{6.7B} & \textbf{13B} & \textbf{Ours (3.65B)} \\
        \midrule
        MultiExpertQA-P (Base)   & 2   & 0.56   & 0.67   & 0.75   & 0.85   & \textbf{0.83}   & 0.21   & 0.26   & 0.32   & 0.55   & 0.59  & \textbf{0.65}  \\
        MultiExpertQA-P (FT)   & 2   & 0.64   & 0.69   & 0.79   & -   & \textbf{0.83}   & 0.34   & 0.41   & 0.44   & 0.58   & 0.62  & \textbf{0.65}  \\
        MultiExpertQA-All (Base)   & 2   & 0.45   & 0.56   & 0.69   & 0.79   & \textbf{0.78}   & 0.19   & 0.24   & 0.29   & 0.52   & 0.57  & \textbf{0.62}  \\
        MultiExpertQA-All (FT)   & 2   & 0.54   & 0.61   & 0.73   & -   & \textbf{0.78}   & 0.28   & 0.29   & 0.34   & 0.53   & 0.58  & \textbf{0.62}  \\
        \toprule
        & & \multicolumn{5}{c}{\textbf{Llama 2}} & \multicolumn{6}{c}{\textbf{OPT}} \\
        \cmidrule(lr){3-7} \cmidrule(lr){8-13}  
        \textbf{Benchmark} & \textbf{\#Exp} & \textbf{7B} & \textbf{13B} & \textbf{34B} & \textbf{70B} & \textbf{Ours (42B)} & \textbf{125M} & \textbf{350M} & \textbf{1.3B} & \textbf{6.7B} & \textbf{13B} & \textbf{Ours (4B)} \\
        \midrule
        
        MultiExpertQA-P (Base)  & 3  & 0.49  & 0.64  & 0.70  & 0.82  & \textbf{0.81}  & 0.19   & 0.24   & 0.28   & 0.51   & 0.59   & \textbf{0.61}  \\
        MultiExpertQA-P (FT)  & 3  & 0.53  & 0.67  & 0.74  & -  & \textbf{0.81}  & 0.26   & 0.29   & 0.32   & 0.58   & 0.63   & \textbf{0.61}  \\
        MultiExpertQA-All (Base)   & 3   & 0.41   & 0.55   & 0.62   & 0.74   & \textbf{0.76}   & 0.15   & 0.22   & 0.25   & 0.49   & 0.54  & \textbf{0.58}  \\
        MultiExpertQA-All (FT)   & 3   & 0.47   & 0.61   & 0.65   & -   & \textbf{0.76}   & 0.23   & 0.27   & 0.31   & 0.52   & 0.58  & \textbf{0.58}  \\
        \bottomrule
    \end{tabularx}
    \label{tab:overall} 
    
\end{table*}

\subsection{Comparison with Mixture-of-Experts}
\label{subsec:moe}

Mixture-of-Experts (MoE) routes each token to a subset of experts within a single large model. MoEs differ from our framework in that they are jointly pretrained, and the router operates at the token level without modeling any multi-step reasoning dependencies. We now compare ~\name{} with two MoE models; Mixtral-8$\times$7B~\cite{mixtral} and OLMoE~\cite{olmoe} on the MultiExpertQA-All benchmark. As shown in Table~\ref{tab:moe_results}, we observe that ~\name{} outperforms both the MoE models in this benchmark since they do not model the logical dependencies across sub-queries.

\begin{table}[h]
\centering
\begin{tabular}{lc}
\toprule
\textbf{Model} & \textbf{F1 Score} \\
\midrule
OLMoE-1B--7B--0924  & 0.52 \\
Mixtral-8$\times$7B & 0.74 \\
\midrule
Comp-LLM (35B)      & \textbf{0.78} \\
\bottomrule
\end{tabular}
\caption{Comparison with MoE baselines on MultiExpertQA-All benchmark.}
\label{tab:moe_results}
\end{table}

\subsection{Comparison of Routing Methods}

Next, we compare ~\name{} with various methods of routing the input query and highlight our improvements as shown in Table ~\ref{tab:routemtd}. We examine two routing approaches: All Experts routing in which each sub-query is sent to all experts for response aggregation, and Random Routing, where a sub-query is randomly assigned to an expert. We observe that ~\name{} performs better than both these methods since the expert router directs each sub-query to the chosen expert and overall benefits from cross-collaboration. Furthermore, we observe that the advantages of ~\name{} becomes more pronounced as the number of experts increases. This is because in random routing there is a higher chance of routing queries to the wrong expert, while averaging across models when routing to all experts dilutes output quality.


\begin{table}[h]
  \centering
  \footnotesize
  \begin{tabularx}{\columnwidth}{lccP{2cm}} 
    \hline
    \textbf{Network} & \textbf{All Experts} & \textbf{Random} & \textbf{~\name{}} \\
    \hline
    \hline
    \multicolumn{4}{c}{\textit{\# Experts: 2}} \\ 
    \hline
    Llama 2  & 0.76   & 0.71   & \textbf{0.78} \\
    OPT     & 0.60  & 0.53  & \textbf{0.62} \\
    \hline
    \multicolumn{4}{c}{\textit{\# Experts: 3}} \\ 
    \hline
    Llama 2  & 0.67  & 0.56  & \textbf{0.76} \\
    OPT     & 0.42  & 0.39  & \textbf{0.58} \\
    \hline
    
  \end{tabularx}
  \caption{Accuracy of ~\name{} in comparison to various expert routing methods.}
  \label{tab:routemtd}
\end{table}
\subsection{Expert Router Analysis}

\begin{figure}[htb]
  \includegraphics[width=\columnwidth]{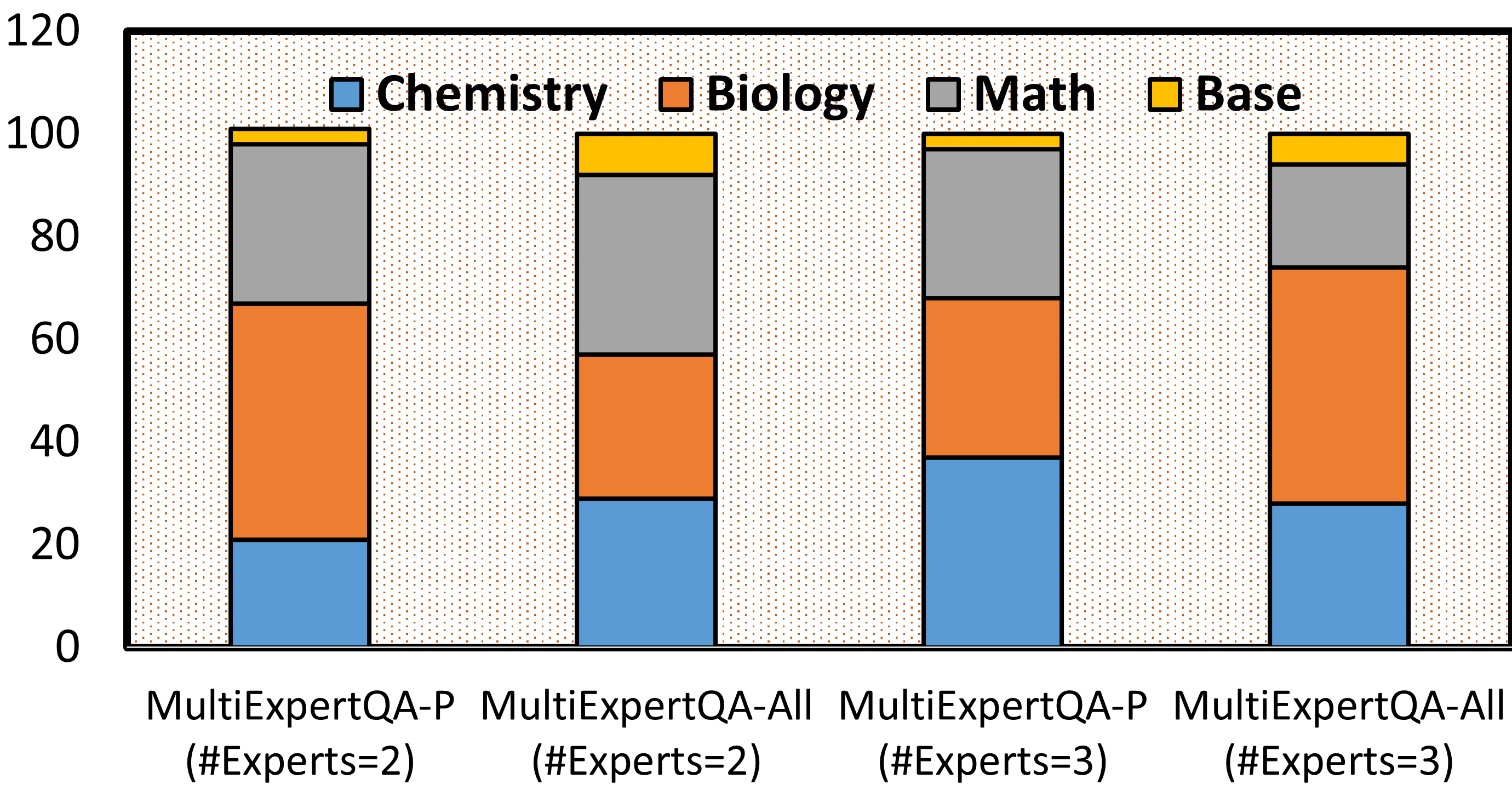}
  \caption{Expert routing distribution for different benchmarks}
  \label{fig:routeres}
  \vspace{-12pt}
\end{figure}

As illustrated in Figure ~\ref{fig:routeres}, we examine how ~\name{} routes its sub-queries across various datasets using the Llama 2 network. Our findings show that most queries are directed to individual domain experts based on their specific expertise. However, some queries that do not meet the threshold requirements are routed to the base model instead. The accuracy improvements can therefore be attributed to the collaborative efforts of the experts in addressing these queries. This collaboration enables a more nuanced approach to complex queries, allowing us to leverage the specialized knowledge of multiple experts as needed.

\subsection{Latency Benefits vs Comp-LLM-seq}

\begin{figure}[htb]
  \includegraphics[width=\columnwidth]{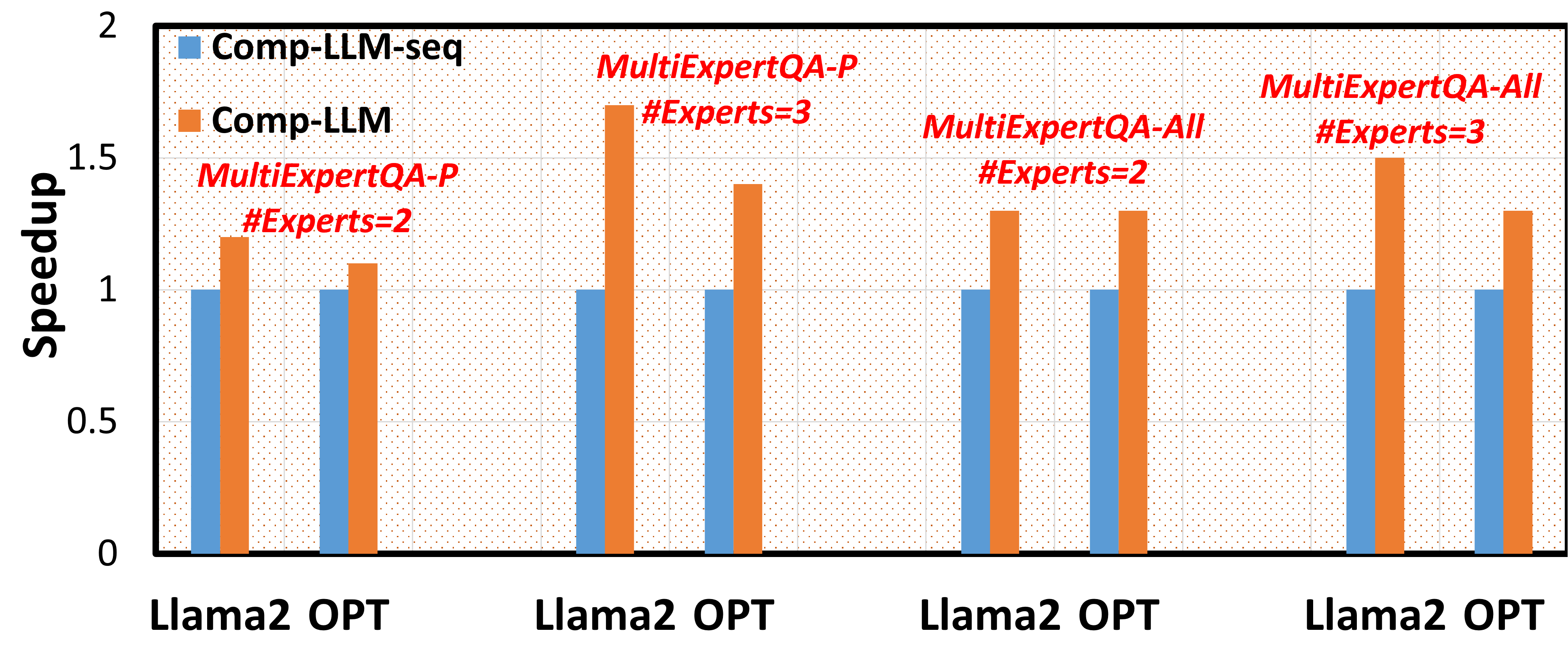}
  \caption{Latency improvements of ~\name{} compared to ~\name{}-seq, which processes sub-queries sequentially}
  \label{fig:lat}
  \vspace{-12pt}
\end{figure}

Figure ~\ref{fig:lat} illustrates the speedup achieved by ~\name{} using the Llama 2 network, which incorporates a runtime scheduler to identify parallel nodes in the query graph, in comparison to ~\name{}-seq, where sub-queries are processed sequentially. We observe that across benchmarks with different number of expert domains, ~\name{} achieves 1.1x-1.7x improvement in latency compared to ~\name{}-seq. In general, we observe that as the number of experts increases, we obtain larger improvements (1.3x-1.7x) due to increase in the number of sub-queries.  We also observe that MultiExpertQA-P (3 experts) shows the largest improvements since none of the sub-queries needs to be processed sequentially due to no dependencies between them.

\subsection{End-to-End Latency Benefits vs Monolithic Models}
\label{subsec:latency}

Although ~\name{} evaluates multiple sub-queries for a single input, the end-to-end latency is lower compared to monolithic LLMs of similar or larger sizes. This is due to the fact that each expert model is significantly smaller than the largest monolithic baselines and independent sub-queries can be executed in parallel by the runtime scheduler. Table~\ref{tab:latency} reports the average inference time per example on the MultiExpertQA-All benchmark using a single NVIDIA A40 GPU. Large monolithic models such as Llama~2--70B exceed single-GPU memory capacity and rely on CPU offloading, resulting in substantial slowdown. In contrast, Comp-LLM executes only the experts required by the query graph, enabling faster execution despite multiple
forward passes.

\begin{table}[h]
\centering
\begin{tabular}{lcc}
\toprule
\textbf{Model} & \textbf{F1 Score} & \textbf{Latency (s)} \\
\midrule
Llama 2 -- 7B   & 0.45 & 4.12 \\
Llama 2 -- 13B  & 0.56 & 4.64 \\
Llama 2 -- 70B  & 0.79 & 892.07 \\
\midrule
Comp-LLM (35B)  & 0.78 & 108.78 \\
\bottomrule
\end{tabular}
\caption{Inference latency on MultiExpertQA-All using a single A40 GPU.
Llama~2--70B requires CPU offloading, which introduces significant slowdown.}
\label{tab:latency}
\end{table}

~\vspace{-12pt}
\section{Conclusion}
\label{sec:conclusion}

In this work, we introduced ~\name{}, a framework designed to enhance the reasoning capabilities of LLMs while reducing their memory footprint. ~\name{} consists of three main components: a Sub-query Generator that outputs a query dependency graph, assigns sub-queries to the appropriate experts and represents their dependencies; a Query Executor with a runtime scheduler to mitigate performance degradation from sequential sub-query processing; and a Response Aggregator that merges expert responses into a coherent answer. In summary, ~\name{} provides more accurate responses compared to state-of-the-art LLMs while achieving lower latency compared to sequentially processing the sub-queries.

\section{Limitations}
\label{sec:limitations}

There are two limitations to our approach. First, \name{} requires that sufficient experts are created to handle all tasks. When none of the experts are suited for a given task, ~\name{} will default to routing it to the base LLM for all sub-queries, which may lead to no improvement in accuracy over using a standard pre-trained LLM with sub-query generation. Second, this work only applies to reasoning tasks that are static in nature. However, for more interactive tasks, such as games, where the user input influences the next decision, the reasoning process becomes dynamic. These tasks require continuous adaptation based on evolving contexts, making our current static sub-query generation approach less effective. This is an interesting problem to be addressed in future work.

\bibliography{custom}

\appendix

\section{List of Hyperparameters}
\clearpage

\begin{table}[h]
  \centering
  \begin{tabular}{cc}
    \hline
    \textbf{Hyperparameter} & \textbf{Value} \\
    \hline
    epochs     & 10  \\
    train\_batch\_size     & 16  \\
    gradient\_accum\_steps & 4 \\
    lr    & 2e-5  \\
    lr\_schedule & cosine \\
    weight\_decay & 0.1 \\
    precision & bf16 \\
    warmup\_ratio & 0.03 \\
    \hline
  \end{tabular}
  \caption{Sub-query Generator.}
  \label{tab:subq_train}
\end{table}

\begin{table}[h]
  \centering
  \begin{tabular}{cc}
    \hline
    \textbf{Hyperparameter} & \textbf{Value} \\
    \hline
    epochs     & 1  \\
    train\_batch\_size     & 8  \\
    gradient\_accum\_steps & 16 \\
    lr    & 2e-5  \\
    lr\_schedule & cosine \\
    weight\_decay & 0.1 \\
    precision & bf16 \\
    warmup\_ratio & 0.04 \\
    \hline
  \end{tabular}
  \caption{OPT experts finetuning}
  \label{tab:opt_exp}
\end{table}

\begin{table}[h]
  \centering
  \begin{tabular}{cc}
    \hline
    \textbf{Hyperparameter} & \textbf{Value} \\
    \hline
    epochs     & 1  \\
    train\_batch\_size     & 8  \\
    gradient\_accum\_steps & 4 \\
    lr    & 2.5e-5  \\
    optim & paged\_adamw\_8bit \\
    weight\_decay & 0.2 \\
    precision & bf16 \\
    warmup\_ratio & 0.04 \\
    \hline
  \end{tabular}
  \caption{Llama 2 experts finetuning}
  \label{tab:llama2_exp}
\end{table}

\label{sec:appendix}

\section{Prompts}

\begin{table*}[h]
\centering

\begin{minipage}{1.9\columnwidth}\vspace{0mm}    \centering
\begin{tcolorbox} 
    \centering
    \small
     \hspace{-6mm}
    \begin{tabular}{p{0.99\columnwidth}}
\begin{minipage}{0.99\columnwidth}\vspace{0mm}
""" \\
You are an intelligent AI assistant and your task is to decompose a complex query into a set of simpler, numbered sub-queries that, when answered, will lead to a comprehensive solution to the original query. Ensure that: \\
1. Each sub-question is as simple as possible, focusing on one key concept or step at a time. \\
2. Number each sub-question sequentially, reflecting the logical flow of reasoning needed to answer the original query. \\
3. Construct a dependency graph by indicating the relationships between the sub-questions. Use their corresponding numbers to show which sub-questions depend on others to be answered first.\\

The sub-questions should be arranged to break down the reasoning into clear, manageable steps, ensuring improved accuracy through multi-step reasoning.\\

Example: Here are some sample query decompositions along with their corresponding dependency graph relations.\\

Input: The gold spike in the city where Falling in Reverse formed is owned by a person whose alma mater has how many undergraduates? \\
Output: \\
1. Where did Falling in Reverse form? \\
2. Who owns the gold spike in that city? \\
3. What is the alma mater of that individual? \\
4. How many undergraduates does that alma mater have? \\
Dependency Graph: "1 -> 2 -> 3 -> 4" \\
Input: Which is greater: Avogadro's number or the GDP of the world's most populous country in the world? \\
Output: \\
1. What is Avogadro's number? \\
2. Which is the most populous country in the world? \\
3. What is the GDP of this country? \\
4. How does these two numbers compare? \\
Dependency Graph: "1 -> 4, 2 -> 3, 3 -> 4" \\
\#\#\# \\
Input: {MultiHop Query} \\
Output: \\
... \\
"""
\end{minipage}
    \end{tabular}
\end{tcolorbox}
\caption{Sub-query Generator Prompt for GPT 4o for generating data}
\label{tab:subq_prompt}
\end{minipage}
\end{table*}

\begin{table*}[htb]\centering

\begin{minipage}{1.9\columnwidth}\vspace{0mm}    \centering
\begin{tcolorbox} 
    \centering
    \small
     \hspace{-6mm}
    \begin{tabular}{p{0.99\columnwidth}}
\begin{minipage}{0.99\columnwidth}\vspace{0mm}
""" \\
You are an intelligent AI assistant, and your task is to combine expert responses such that overall response answers the query. Your answer must be coherent and should answer the original query by using the expert responses as context. Please provide a detailed, well-structured and error free answer to the original query. \\
Query: \{\}\\
The expert responses are given below: \\
Response from Expert 1: \{\} \\
Response from Expert 2: \{\} \\
Response from Expert 3: ... \\
\#\#\# \\
Output: \\
... \\
"""
\end{minipage}
    \end{tabular}
\end{tcolorbox}
\caption{Response Aggregator Prompt for GPT 4o.}
\label{tab:response_prompt}
\end{minipage}
\end{table*}

\begin{table*}[htb]\centering

\begin{minipage}{1.9\columnwidth}\vspace{0mm}    \centering
\begin{tcolorbox} 
    \centering
    \small
     \hspace{-6mm}
    \begin{tabular}{p{0.99\columnwidth}}
\begin{minipage}{0.99\columnwidth}\vspace{0mm}
""" \\
You are an intelligent AI assistant, and your task is to combine the two questions into a multi-hop question. 
Ensure the combined multihop question is clear and unambiguous.\\
Question 1: \{\}\\
Question 2: \{\}\\
\#\#\# \\
Multihop Question: \\
... \\
"""
\end{minipage}
    \end{tabular}
\end{tcolorbox}
\caption{MultiExpertQA-P prompt for GPT 4o.}
\label{tab:multiexpqap_prompt}
\end{minipage}
\end{table*}

\begin{table*}[htb]\centering

\begin{minipage}{1.9\columnwidth}\vspace{0mm}    \centering
\begin{tcolorbox} 
    \centering
    \small
     \hspace{-6mm}
    \begin{tabular}{p{0.99\columnwidth}}
\begin{minipage}{0.99\columnwidth}\vspace{0mm}
""" \\
You are an intelligent AI assistant, and your task is to process the given input through the following steps: \\
1. Use Named Entity Recognition (NER) to identify entities in the input. \\
2. Select two of the identified entities. \\
3. For each of these two entities: 
    \begin{itemize}
    \item Generate a concise \{Expert\}-related fact.
    \item Transform this fact into a question where the entity is the answer.
    \item Ensure the entity itself is not mentioned in the question.
    \end{itemize}
4. Create a multi-hop question by replacing the original mentions of these two entities in the input with their respective questions. \\
5. Provide an answer to the resulting multi-hop question.\\
\#\#\# \\
Please present your results as follows: \\
1. Original input \\
2. List of identified entities \\
3. For each of the two selected entities: 
    \begin{itemize}
    \item Entity name 
    \item Injected fact 
    \item  Corresponding question 
    \end{itemize}
4. New multi-hop question \\
5. Answer to the multi-hop question \\
Maintain accuracy in all facts and ensure questions are clear and unambiguous.\\
"""
\end{minipage}
    \end{tabular}
\end{tcolorbox}
\caption{MultiExpertQA-All prompt for GPT 4o.}
\label{tab:multiexpqaall_prompt}
\end{minipage}
\end{table*}
\label{sec:appendix1}

\end{document}